\documentclass[11pt]{article}

\usepackage[final]{acl}

\usepackage{times}
\usepackage{latexsym}

\usepackage[T1]{fontenc}

\usepackage[utf8]{inputenc}

\usepackage{microtype}

\usepackage{inconsolata}

\usepackage{graphicx}
\usepackage{amsmath,amssymb}
\usepackage{algorithm}
\usepackage{algpseudocode} 
\usepackage{booktabs,multirow}
\usepackage{booktabs}
\usepackage{multirow}
\usepackage{array}
\usepackage[table]{xcolor}
\definecolor{decgreen}{RGB}{0,128,0} 
\newcommand{\dec}[1]{\textcolor{decgreen}{(#1)}}
\usepackage{subcaption}
\usepackage{enumitem}
\usepackage{marvosym}
\title{GRASPrune: Global Gating for Budgeted Structured Pruning of Large Language Models}

\author{
  \textbf{Ziyang Wang}\textsuperscript{1}\quad
  \textbf{Jiangfeng Xiao}\textsuperscript{2}\quad
  \textbf{Chuan Xiao}\textsuperscript{3}\quad
  \textbf{Ruoxiang Li}\textsuperscript{2}\quad
  \textbf{Rui Mao}\textsuperscript{2}\quad
  \textbf{Jianbin Qin}\textsuperscript{2,\Letter}
  \\
  \textsuperscript{1}Beijing Institute of Technology, Zhuhai\quad
  \textsuperscript{2}Shenzhen University\quad
  \textsuperscript{3}Osaka University
  \\
  \texttt{ziyangwang@bitzh.edu.cn}\quad
  \texttt{2210273035@email.szu.edu.cn}\quad
  \texttt{chuanx@ist.osaka-u.ac.jp}
  \\
  \texttt{liruoxiang@szu.edu.cn}\quad
  \texttt{mao@szu.edu.cn}\quad
  \texttt{qinjianbin@szu.edu.cn}
  \\ 
}

\begin{document}

\pagestyle{plain}
\pagenumbering{arabic}

\maketitle

\begingroup
\renewcommand\thefootnote{}
\footnotetext{Corresponding author: \texttt{qinjianbin@szu.edu.cn}}
\endgroup

\begin{abstract}
Large language models (LLMs) are expensive to serve because model parameters, attention computation, and KV caches impose substantial memory and latency costs. We present GRASPrune, a structured pruning framework applied after pretraining that jointly prunes FFN channels and KV head groups under a single global budget. Instead of learning importance scores without constraints and applying the budget only after training, GRASPrune learns lightweight gate scores with a projected straight-through estimator that enforces a hard mask satisfying the budget at every step while keeping the backbone weights frozen. After the mask is fixed, we calibrate scaling factors on the retained units to mitigate scale mismatch caused by pruning, and fold these factors into the pruned weights to obtain a smaller dense checkpoint with no extra parameters at inference. On LLaMA-2-7B, GRASPrune removes 50\% of parameters and achieves 12.18 perplexity on WikiText-2 while maintaining competitive average zero-shot accuracy on five benchmarks, using four epochs on 512 unlabeled calibration sequences on a single NVIDIA A100~80GB GPU without any full model fine-tuning.\footnote{ \url{https://github.com/ZiY-Wang/GRASPrune}}

\end{abstract}

\section{Introduction}

\begin{figure}[t]
    \centering
    \includegraphics[width=\linewidth]{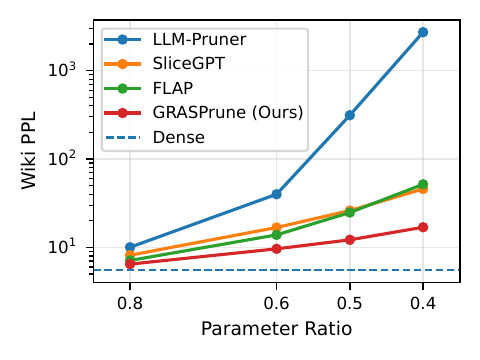}
    \caption{WikiText-2 perplexity of LLaMA-2-7B under different parameter retention ratios.}
    \label{fig:PTB_ppl}
\end{figure}

Large language models (LLMs) achieve strong performance in reasoning and generation~\citep{vaswani2017attention,brown2020language,openai2023gpt4}, but serving them remains expensive. In practice, both large parameter counts and the KV cache that grows with sequence length contribute substantially to memory use and latency~\citep{kaddour2023challenges}. This motivates structural compression methods that reduce checkpoint size and cache-related memory.

Beyond compression, runtime system techniques can also reduce serving cost without modifying model parameters by alleviating bottlenecks in attention and KV caching. Examples include sparse attention kernels~\citep{jiang2024minference,xiao2024infllm} and methods that manage or compress the KV cache~\citep{zhang2023h2o,li2024snapkv,xiao2024efficient}. These techniques target runtime attention and memory traffic and are largely complementary to structural compression.

In this work, we study structured pruning after pretraining. By removing channels or head groups, pruning produces a smaller dense checkpoint that can be deployed with standard serving stacks. As shown in Figure~\ref{fig:PTB_ppl}, model quality degrades nonlinearly as the pruning budget becomes tighter, which suggests that uniform sparsity is insufficient and that the budget must be allocated in a principled way that accounts for unit cost.

Under deployment constraints, structured pruning becomes a discrete selection problem over heterogeneous structure types. Existing pipelines often fall short in three ways. First, FFN channels and KV head groups are usually pruned with separate criteria, even though they compete for the same deployment budget and representational capacity~\citep{li2024lorap,he2025olica}. Second, many methods prescribe keep ratios for each layer or use depth dependent schedules~\citep{tao2023structured,ma2023llm}, which hard code how the budget is distributed across layers and structures instead of learning a single global allocation. Third, because the final selection is discrete and unit costs differ, many pipelines estimate importance scores first and impose the budget only afterward, often followed by reconstruction or additional fine-tuning~\citep{an2024fluctuation,meng2024osscar}. This weakens the connection between the final sparse architecture and the training objective, and it increases engineering overhead. Although a few recent works take a more global view~\citep{kurtic2023ziplm,gao2024disp}, they often ignore cost differences between FFN and KV structures or rely on expensive search and reconstruction.

Our key observation is that the main issue is not the lack of another saliency metric, but the mismatch between how scores are learned and how the final mask is selected. Existing score first and select later pipelines learn importance under an unconstrained surrogate and impose the deployment budget only at the end. We instead enforce budget feasibility inside the optimization loop, so gate scores are learned under the same constraint that defines the final deployable mask.

Based on this view, we propose GRASPrune, a structured pruning framework with a single resource budget. The backbone weights remain frozen, and we optimize only a small set of scalar gates on a small unlabeled calibration set. GRASPrune jointly selects FFN channels and KV head groups under one global budget and compiles the resulting mask into a smaller dense checkpoint that can be served with the standard stack. To match prior pruning baselines, we use parameter count as the budget proxy, and we also report KV cache and throughput results to connect pruning to serving performance. On LLaMA-2-7B, our pruned model reaches 12.18 perplexity on WikiText-2 at a 50\% parameter ratio and 16.65 at 40\%. Learning this allocation is also efficient: four epochs of gate learning plus a brief calibration stage take roughly 6 minutes on a single A100 80GB GPU.

Our contributions are as follows:

\begin{itemize}[leftmargin=*]
\item \textbf{A single global pruning formulation over FFN channels and KV head groups.}
We formulate structured pruning as one global selection problem over FFN channels and KV head groups with different unit costs, and optimize a single mask under one budget constraint.

\item \textbf{Projected STE training with budget feasibility enforced at every step.}
We argue that scores should be learned while the budget is active during training. If learned scores are post-hoc re-ranked by a cost-normalized criterion, the resulting allocation can shift toward cheaper units and degrade performance in our setting. Motivated by this, we use a projected STE loop that applies a hard mask satisfying the budget at each step while optimizing only scalar gates.

\item \textbf{A smaller dense checkpoint through calibration on the retained units.}
After selecting the retained units, we calibrate per unit scales only on this support and fold them into the sliced weights, producing a smaller dense checkpoint without adding extra parameters during inference.
\end{itemize}

\begin{figure*}[t] 
  \centering
  \includegraphics[width=\textwidth]{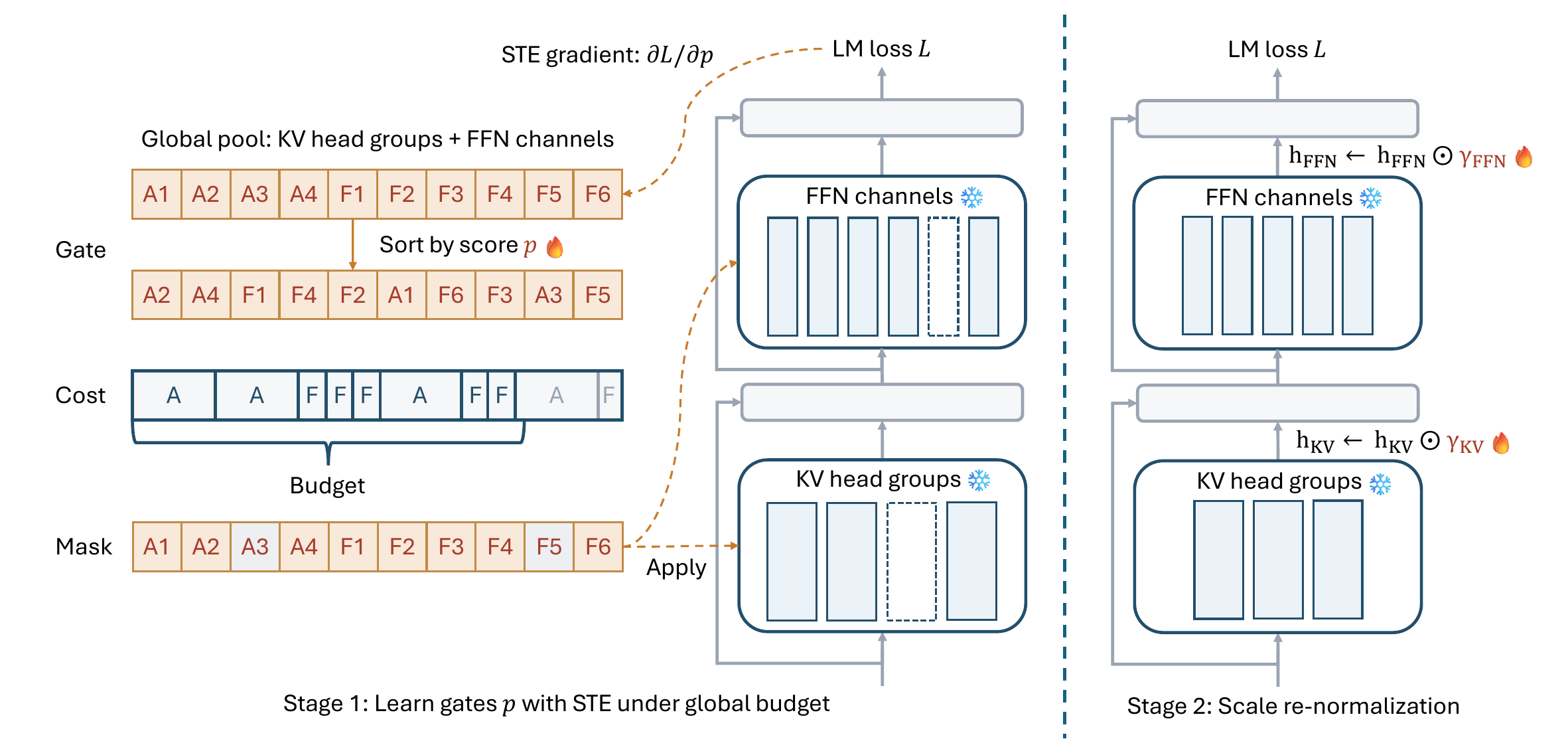}
  \caption{Overall pipeline of GRASPrune.}
  \label{fig:pipeline}
\end{figure*}

\section{Related Work}

\subsection{Pruning for LLMs}

We focus on structured pruning, which removes channels, heads, or layers. Existing LLM-oriented methods can be roughly grouped into three families by how they allocate sparsity. The first family uses fixed per-layer sparsity schedules with local saliency inside each layer: given a target global sparsity, they pre-assign layer-wise keep ratios and then prune the least important structures according to magnitude- or activation-based scores, sometimes followed by light calibration or fine-tuning~\citep{sun2023simple,yang2025wanda++,hu2025fasp}. LLM-Pruner is a representative example that ranks structures with Taylor-style gradients under an OWL-style per-layer schedule~\citep{ma2023llm}.

A second family keeps the layer-wise view but lets data or error decide the ratios, deriving layer-wise sparsity from activations or reconstruction loss instead of prescribing it. SlimGPT is a prominent instance of this design: it extends Optimal Brain Surgeon to structured pruning, using batched greedy updates together with an incremental pruning schedule that is explicitly tuned to layer-wise error, so that more sensitive layers are pruned less aggressively~\citep{ling2024slimgpt}.

A third family adopts a genuinely global view, coordinating sparsity across layers and structure types under a single constraint. ZipLM aggregates multi-granularity saliency into a global ordering and selects structures directly under a target sparsity~\citep{kurtic2023ziplm}. DISP-LLM further relaxes the coupling along the embedding dimension so that different blocks can use different subsets of feature maps and widths, effectively turning pruning into a dimension-independent architecture search while still operating without additional parameters~\citep{gao2024disp}.

\paragraph{Concurrent work.}
Concurrently with our work, E$^3$-Pruner~\citep{yuan2025e3prunerefficienteconomicaleffective} independently propose a related structured pruning framework for LLMs based on differentiable masks. While both methods learn discrete structure selections, their approach focuses on layer-wise depth pruning with knowledge distillation for performance recovery, whereas we target global budgeted FFN--KV pruning at a finer granularity and do not include any explicit distillation-based recovery stage. We regard these lines as complementary evidence that differentiable structured pruning is a promising direction for LLM compression.

\subsection{STE and Gate-Based Sparsification}

A common way to train discrete selections is to place continuous-valued gates on prunable units and pass gradients through the binary hardening step with a straight-through estimator (STE)~\citep{bengio2013estimating,courbariaux2015binaryconnect}. L$_0$ / hard-concrete approaches penalize the expected number of active gates~\citep{louizos2017learning,maddison2016concrete,jang2016categorical}, but under strict deployment budgets stochastic gates are cost-agnostic and can be unstable.

We follow the standard STE paradigm for training discrete gate selections, but unlike stochastic/penalty-based gates that can be cost-agnostic under strict deployment budgets, our method enforces feasibility through an explicit budget projection and uses STE only as a gradient surrogate for updating the underlying gate scores.

\section{Proposed Method}
\label{sec:method}

We propose a global budgeted structured pruning framework that learns which FFN
channels and KV head groups to keep under a single resource budget.
At each iteration, we project gate probabilities to a budget-feasible hard mask for the forward pass, and update the gate scores via an STE surrogate gradient while keeping the backbone frozen.

\subsection{Global Budgeted Structured Pruning}
\label{sec:global-budget}

We consider a pretrained Transformer with parameters $\theta$.
In each layer, the hidden size is $d$, there are $H$ query heads and
$H^{\mathrm{kv}}$ key--value heads, and the head dimension $d_h$
satisfies $d = H d_h$.
We prune two types of structures:
(i) FFN intermediate channels, and
(ii) KV head groups. Under GQA, one prunable KV unit in our implementation corresponds to one KV head group, which is a shared KV structure serving a query group.

Let $\mathcal{S}$ be the set of all prunable units (FFN channels and
KV head groups) across layers.
For each $i \in \mathcal{S}$, a binary variable $z_i \in \{0,1\}$
indicates whether unit $i$ is kept.
We write $\mathbf{z}$ for the concatenation of all $z_i$.
Each unit carries a nonnegative cost $c_i \ge 0$ that approximates
its contribution to parameters.
Given a target keep ratio $\rho \in (0,1]$, we define a single global
budget $B = \rho \sum_{i \in \mathcal{S}} c_i$.

Structured pruning is then formulated as a budget-constrained
optimization problem:
\begin{equation}
\min_{\mathbf{z} \in \{0,1\}^{|\mathcal{S}|}}
\;\mathcal{L}(\theta;\mathbf{z})
\quad
\text{s.t.}\quad
\sum_{i \in \mathcal{S}} c_i z_i \le B,
\label{eq:main}
\end{equation}
where $\mathcal{L}$ is the language-modeling loss of the masked model
$f(x;\theta,\mathbf{z})$ on the data distribution.
In our setting, $\theta$ is fixed and we only learn the structure
$\mathbf{z}$, turning structured pruning into a post-hoc optimization
over the discrete mask while keeping the pretrained backbone frozen.

Since FFN channels and KV head groups have different parameter footprints, we adopt a
simple two-level cost model: each FFN intermediate channel has unit cost
$c_i = 1$, while each KV head group has cost $c_i = \alpha$.
Letting $G = H / H^{\mathrm{kv}}$, we set
\begin{equation}
\alpha = \frac{(2G + 2) d_h}{3},
\label{eq:alpha}
\end{equation}
a parameter-count approximation that puts KV projections under GQA on the same
scale as the three FFN projections per channel. We compute $\alpha$ once from
the configuration and share it across layers so that FFN channels and KV head groups
compete under the same global budget~$B$.

We use this parameter-based proxy mainly to align with common baselines under
matched structural budgets. However, the cost model is flexible: $\alpha$ can be
redefined to target deployment metrics. For instance, under a fixed maximum
generation length, KV-head pruning also reduces KV-cache memory/bandwidth, which
can be incorporated into $\alpha$ in addition to parameter memory.

In practice, we find the method is not overly sensitive to moderate
misspecification of $\alpha$; Appendix~\ref{sec:sensitivity_alpha}
reports an ablation that perturbs $\alpha$ and shows only minor changes.

\subsection{Differentiable Global Mask Optimization}
\label{sec:diff-mask}

Given the global budgeted formulation in Section~\ref{sec:global-budget},
we now describe how to optimize the pruning pattern. A natural pipeline is
to first learn unconstrained importance scores for all units and then solve
a budgeted selection problem only once at the end. Under heterogeneous unit
costs, however, this decouples score learning from the constraint that
defines the final deployable mask: scores are learned without budget
pressure, while costs are imposed only at selection time. We instead
enforce budget feasibility inside the optimization loop, so gate scores are
learned under the same constraint that determines the final mask.

We introduce a continuous score for each prunable unit and alternate
between two steps: (i) projecting the current scores to a hard mask that
satisfies the global budget, and (ii) updating the scores with a
straight-through estimator (STE), while keeping all backbone projections
frozen.

\paragraph{Per-step budget-feasible projection.}
At each training step, we convert relaxed gate probabilities
$\mathbf{p}=(p_i)_{i\in\mathcal{S}}$ into a hard mask
$\mathbf{m}\in\{0,1\}^{|\mathcal{S}|}$ under heterogeneous unit costs
$\mathbf{c}=(c_i)_{i\in\mathcal{S}}$:
\begin{equation}
\mathbf{m}=\textsc{Project}(\mathbf{p},\mathbf{c},B).
\label{eq:project}
\end{equation}

$\textsc{Project}$ sorts all units by $p_i$ in descending order, scans this
order once, and sets $m_i=1$ until adding the next unit would exceed the
budget. In this way, $p_i$ determines a global ordering of utility, while
cost is used only to enforce feasibility.

This distinction is important. Since $\mathbf{p}$ is updated under repeated budget-feasible projections, the learned scores are adapted to the same selection rule used during training. A natural
alternative is to rank units by a cost-normalized criterion such as
$p_i/c_i$. In our setting, however, this changes the selection rule used after gate learning and empirically shifts the allocation toward cheaper units. We therefore rank by $p_i$ and use
cost only in the projection step. This effect is validated by the $p/c$
ablation in Table~\ref{tab:pc_ablation}.

Operationally, \textsc{Project} implements a feasible prefix selection rule under the ordering induced by the learned probabilities $\mathbf{p}$.
This projection requires one global sort and one linear scan, giving an
$O(n \log n)$ cost per step. In practice, however, the overhead is negligible
compared with the Transformer forward/backward pass. On a single A100 GPU
using \texttt{torch.argsort}, the cumulative sorting time accounts for only
0.11\% of the total gate-learning time on LLaMA-2-7B and 0.14\% on
LLaMA-2-13B. We provide the detailed wall-clock measurements in Appendix~\ref{app:wall_clock_proj}.

To avoid degenerate layers, if a layer would prune all FFN channels or all
KV heads, we set the highest-$p_i$ unit in that group to $1$.

\paragraph{Hard forward, soft backward.}
Once the hard mask $\mathbf{m}$ is constructed, we use it in the forward
pass and back-propagate through the soft probabilities $\mathbf{p}$. For
each unit, we define a surrogate gate
\begin{equation}
\tilde{z}_i
=
m_i + \bigl(p_i - \mathrm{stopgrad}(p_i)\bigr),
\label{eq:ste}
\end{equation}
where $m_i \in \{0,1\}$ is the hard mask and
$\mathrm{stopgrad}(\cdot)$ denotes the stop-gradient operator. In the
forward pass, the model sees $\tilde z_i = m_i$ and therefore behaves as if
a deterministic binary mask were applied. In the backward pass, the
gradient flows through $p_i$ as if $\tilde z_i = p_i$, yielding
\begin{equation}
\frac{\partial \mathcal{L}}{\partial s_i}
=
\frac{\partial \mathcal{L}}{\partial \tilde{z}_i}
\cdot
\frac{\partial p_i}{\partial s_i}
=
\frac{1}{\tau}
\frac{\partial \mathcal{L}}{\partial \tilde{z}_i}
\, p_i (1-p_i),
\label{eq:grad}
\end{equation}
which is the standard STE surrogate for Bernoulli gates. We update the
scores $\mathbf{s}$ with AdamW on this surrogate gradient while keeping
$\theta$ fixed.

\paragraph{Optimization view.}
Let $\hat{\mathbf{m}}(\mathbf{p})$ denote the hard mask returned by $\textsc{Project}(\mathbf{p},\mathbf{c},B)$. The corresponding hard-mask objective can be written as
\begin{equation}
\tilde{\mathcal{L}}(\mathbf{s})
=
\mathbb{E}_{x}\big[
L\big(f(x;\theta,\hat{\mathbf{m}}(\mathbf{p}(\mathbf{s})))\big)
\big].
\label{eq:true}
\end{equation}

This objective is non-smooth because $\hat{\mathbf{m}}$ is piecewise constant in $\mathbf{p}$. STE handles this discontinuity by using the hard mask in the forward pass and propagating gradients through the relaxed probabilities in the backward pass. Over
training, units with consistently strong learning signals accumulate higher
scores and are repeatedly selected, whereas weak or noisy units are pushed
toward $p_i \approx 0$. This interaction encourages score polarization and
yields a stable sparse mask under the prescribed budget.

\paragraph{Budget-preserving scaling calibration.}
Prior work observes that structured changes such as head pruning can shift the scale of attention
outputs and that lightweight rescaling of the remaining components can mitigate this effect~\citep{wong2025a3}.
Motivated by this, we add a post-hoc calibration step that does not change the sparsity pattern.

After the final projection $\textsc{Project}(\mathbf{p},\mathbf{c},B)$, we freeze the hard mask $\mathbf{m}$ and
introduce a scalar multiplier $\gamma_i$ for each retained unit
$\mathcal{I}=\{i\mid m_i=1\}$, applied multiplicatively to the corresponding channel/head output.
We optimize only $\{\gamma_i\}_{i\in\mathcal{I}}$ on the same calibration set with the backbone weights fixed,
so the number of trainable parameters is $O(|\mathcal{I}|)$ and the FLOPs are unchanged.
After calibration, we materialize the pruned model by slicing FFN and attention projections according to
$\mathbf{m}$ and folding $\gamma$ into the sliced weights as a diagonal reweighting on the retained support.
Finally, we discard all gate and scale parameters, yielding a smaller dense checkpoint with no extra
inference-time overhead.

\section{Experiments}

\begin{table*}[t]
\centering
\small
\renewcommand{\arraystretch}{1.25}
\setlength{\tabcolsep}{2.5pt}

\begin{tabular}{ll|ccc|cccccc}
\toprule
\textbf{Param Ratio} & \textbf{Method} & \textbf{Wiki} & \textbf{PTB} & \textbf{C4} 
& \textbf{ARC-C} & \textbf{ARC-E} & \textbf{HellaSwag} & \textbf{PIQA} & \textbf{WinoGrande} & \textbf{Avg.} \\
\midrule
1.0 & Original & 5.4911 & 27.028 & 7.2745 & 0.4616 & 0.7487 & 0.7597 & 0.7873 & 0.6906 & 0.6896 \\
\midrule
\multirow{4}{*}{0.8}
 & LLM-Pruner & 10.0406 & 53.1254 & 12.2540 & 0.3703 & \textbf{0.6393} & \underline{0.6666} & \textbf{0.7486} & 0.6243 & \underline{0.6098} \\
 & SliceGPT   & 8.1582 & 95.0674 & 38.2862 & 0.3336 & 0.5206 & 0.5111 & 0.6534 & 0.6251 & 0.5288 \\
 & FLAP       & \underline{7.1198} & \textbf{36.9429} & \textbf{10.7090} & \underline{0.3720} & 0.6166 & 0.6545 & \underline{0.7465} & \textbf{0.6346} & 0.6048 \\
 \rowcolor[gray]{0.9}
 \cellcolor{white} & GRASPrune  & \textbf{6.4700} & \underline{48.1827} & \underline{11.4444} & \textbf{0.3848} & \underline{0.6351} & \textbf{0.6748} & 0.7405 & \textbf{0.6346} & \textbf{0.6139} \\
\midrule
\multirow{4}{*}{0.6}
 & LLM-Pruner & 39.9448 & 207.6673 & 34.0334 & 0.2747 & \underline{0.4386} & 0.4215 & 0.6442 & 0.5375 & 0.4633 \\
 & SliceGPT   & 16.7937 & 252.5093 & 107.1938 & 0.2543 & 0.3796 & 0.3504 & 0.5555 & 0.5446 & 0.4169 \\
     & FLAP       & \underline{13.8587} & \underline{82.2823} & \underline{20.2035} & \underline{0.2875} & 0.3990 & \underline{0.5052} & \textbf{0.6757} & \textbf{0.5714} & \underline{0.4878} \\
 \rowcolor[gray]{0.9}
 \cellcolor{white} & GRASPrune  & \textbf{9.6371} & \textbf{70.1054} & \textbf{18.8686} & \textbf{0.2995} & \textbf{0.4899} & \textbf{0.5092} & \underline{0.6523} & \underline{0.5620} & \textbf{0.5026} \\
\midrule
\multirow{4}{*}{0.5}
 & LLM-Pruner & 312.9657 & 473.505 & 279.447 & 0.2526 & 0.2950 & 0.2960 & 0.5609 & 0.5304 & 0.3870 \\
 & SliceGPT   & 26.1031 & 365.6156 & 175.993 & 0.2406 & 0.3270 & 0.3114 & 0.5234 & 0.5241 & 0.3853 \\
 & FLAP       & \underline{24.8023} & \underline{198.9331} & \underline{36.5839} & \underline{0.2688} & \underline{0.3750} & \underline{0.4115} & \underline{0.6224} & \textbf{0.5470} & \underline{0.4449} \\
 \rowcolor[gray]{0.9}
 \cellcolor{white} & GRASPrune  & \textbf{12.1824} & \textbf{123.0390} & \textbf{27.8860} & \textbf{0.2841} & \textbf{0.4343} & \textbf{0.4377} & \textbf{0.6279} & \underline{0.5430} & \textbf{0.4654} \\
\midrule
\multirow{4}{*}{0.4}
 & LLM-Pruner & 2724.816 & 1917.148 & 998.496 & \textbf{0.2543} & 0.2778 & 0.2667 & 0.5365 & 0.4909 & 0.3652 \\
 & SliceGPT   & \underline{45.8675} & 526.8949 & 358.6953 & 0.2415 & 0.2824 & 0.2869 & 0.5136 & \underline{0.5020} & 0.3653 \\
 & FLAP       & 51.6924 & \underline{426.1057} & \underline{95.4491} & \underline{0.2500} & \underline{0.3657} & \underline{0.3090} & \underline{0.5642} & 0.4933 & \underline{0.3964} \\
 \rowcolor[gray]{0.9}
 \cellcolor{white} & GRASPrune  & \textbf{16.6515} & \textbf{148.4131} & \textbf{43.1906} & 0.2389 & \textbf{0.3704} & \textbf{0.3375} & \textbf{0.5664} & \textbf{0.5170} & \textbf{0.4060} \\
\bottomrule
\end{tabular}
\caption{
Comprehensive evaluation of pruning methods on LLaMA-2-7B.
We report perplexity (WikiText-2, PTB, and C4) and zero-shot accuracy on five downstream benchmarks. 
Lower PPL and higher accuracy indicate better performance.
}

\label{tab:llama2_7b_full}
\end{table*}

\begin{table*}[t]
\centering
\small
\renewcommand{\arraystretch}{1.25}
\setlength{\tabcolsep}{2pt}

\begin{tabular}{ll|ccc|cccccc}
\toprule
\textbf{Model} & \textbf{Method} & \textbf{Wiki} & \textbf{PTB} & \textbf{C4} 
& \textbf{ARC-C} & \textbf{ARC-E} & \textbf{HellaSwag} & \textbf{PIQA} & \textbf{WinoGrande} & \textbf{Avg.} \\
\midrule
\multirow{4}{*}{LLaMA-7B}
 & LLM-Pruner & 3223.184 & 2258.9519 & 1223.3826 & \underline{0.2645} & \underline{0.2891} & 0.2766 & 0.5326 & \underline{0.5162} & 0.3758 \\
 & SliceGPT   &   \underline{56.7939} &  \underline{417.2273} &  298.6589 & 0.2423 & 0.2664 & 0.2637 & 0.5071 & 0.4846 & 0.3528 \\
 & FLAP       &  104.8303 &  586.9854 &  \underline{237.1635} & \textbf{0.2688} & 0.2799 & \underline{0.3019} & \underline{0.5522} & 0.5099 & \underline{0.3825} \\
 \rowcolor[gray]{0.9}
 \cellcolor{white} & GRASPrune  &  \textbf{15.8889} &   \textbf{49.7172} &   \textbf{40.5738} & 0.2457 & \textbf{0.3531} & \textbf{0.3442} & \textbf{0.5734} & \textbf{0.5209} & \textbf{0.4074} \\
\midrule
\multirow{4}{*}{LLaMA-2-7B}
 & LLM-Pruner & 2724.8161 & 1917.1476 & 998.4958 & \textbf{0.2543} & 0.2778 & 0.2667 & 0.5365 & 0.4909 & 0.3654 \\
 & SliceGPT   &   \underline{45.8675} &  526.8949 & 358.6953 & 0.2415 & 0.2824 & 0.2869 & 0.5136 & \underline{0.5020} & 0.3652 \\
 & FLAP       &   51.6924 &  \underline{426.1057} &  \underline{95.4491} & \underline{0.2500} & \underline{0.3657} & \underline{0.3090} & \underline{0.5642} & 0.4933 & \underline{0.3964} \\
 \rowcolor[gray]{0.9}
 \cellcolor{white} & GRASPrune  &   \textbf{16.6515} & \textbf{148.4131} & \textbf{43.1906} & 0.2389 & \textbf{0.3704} & \textbf{0.3375} & \textbf{0.5664} & \textbf{0.5170} & \textbf{0.4060} \\
\midrule
\multirow{4}{*}{Vicuna-7B}
 & LLM-Pruner & 562.2976 & 1685.2808 & 374.5709 & \underline{0.2619} & 0.3077 & \underline{0.2904} & 0.5495 & 0.5051 & 0.3829 \\
 & SliceGPT   &  \underline{51.4984} &  \underline{523.8676} & \underline{328.0323} & 0.2329 & 0.2992 & 0.2873 & 0.5163 & 0.5083 & 0.3688 \\
 & FLAP       &  90.7231 &  575.6321 & 421.1414 & 0.2551 & \underline{0.3582} & 0.2740 & \underline{0.5506} & \underline{0.5114} & \underline{0.3899} \\
 \rowcolor[gray]{0.9}
 \cellcolor{white} & GRASPrune  &  \textbf{17.1801} &  \textbf{157.9849} &  \textbf{51.2902} & \textbf{0.2628} & \textbf{0.3805} & \textbf{0.3514} & \textbf{0.5734} & \textbf{0.5122} & \textbf{0.4161} \\
\midrule
\multirow{4}{*}{LLaMA-2-13B}
 & LLM-Pruner & 919.8574 & 1067.0533 & 693.5800 & 0.2466 & 0.3043 & 0.3026 & 0.5691 & \underline{0.5178} & 0.3881 \\
 & SliceGPT   &  35.8700 &  \underline{475.6130} & 150.9691 & 0.2304 & 0.3043 & 0.2959 & 0.5218 & 0.5099 & 0.3725 \\
 & FLAP       &  \underline{29.8589} &  509.9817 &  \underline{42.4381} & \textbf{0.2722} & \textbf{0.4057} & \underline{0.3818} & \underline{0.5947} & \underline{0.5178} & \underline{0.4344} \\
 \rowcolor[gray]{0.9}
 \cellcolor{white} & GRASPrune  &  \textbf{12.1824} &  \textbf{153.1242} &  \textbf{32.0965} & \underline{0.2679} & \underline{0.3994} & \textbf{0.4032} & \textbf{0.6066} & \textbf{0.5288} & \textbf{0.4412} \\
\bottomrule
\end{tabular}
\caption{Cross-model comparison of pruning methods at 40\% parameter retention on LLaMA-7B, LLaMA-2-7B, Vicuna-7B, and LLaMA-2-13B.}
\label{tab:multi-model-main}
\end{table*}

\begin{table*}[t]
\centering
\small
\setlength{\tabcolsep}{3pt} 

\begin{tabular}{c c cc cc}
\toprule
\multirow{2}{*}{\textbf{Param Ratio}} &
\multirow{2}{*}{\textbf{Weights (MiB)}} &
\multicolumn{2}{c}{\textbf{$T=256$}} &
\multicolumn{2}{c}{\textbf{$T=2048$}} \\
\cmidrule(lr){3-4}\cmidrule(lr){5-6}
& & \textbf{KV (MiB)} & \textbf{Peak (MiB)}
  & \textbf{KV (MiB)} & \textbf{Peak (MiB)} \\
\midrule
1.0 & 12852.5
    & 512.0  & 13990.8
    & 4096.0 & 21889.8 \\
0.8 & 10382.1\,\dec{-19.2\%}
    & 418.5\,\dec{-18.3\%}  & 11340.6\,\dec{-18.9\%}
    & 3348.0\,\dec{-18.3\%} & 17822.5\,\dec{-18.6\%} \\
0.6 &  7910.7\,\dec{-38.4\%}
    & 303.0\,\dec{-40.8\%}  &  8644.5\,\dec{-38.2\%}
    & 2424.0\,\dec{-40.8\%} & 13425.8\,\dec{-38.7\%} \\
\bottomrule
\end{tabular}
\caption{Memory breakdown under different prefill lengths $T$ ($B=4$). Weights memory is independent of $T$; KV-cache and peak scale with $T$. Full results are provided in Appendix \ref{sec:mem_breakdown_full}.}
\label{tab:mem_breakdown_compact}
\end{table*}

\subsection{Experimental Settings}

\paragraph{Models and evaluation protocols.}
Following the previous works \citep{ma2023llm,li2024lorap,he2025olica}, we mainly focus on the evaluation of LLaMA \cite{touvron2023llama}, LLaMA-2 \cite{touvron2023llama2}, Vicuna \cite{vicuna2023} models. For generation task, we evaluate the model’s perplexity on WikiText-2 \cite{merity2016pointer}, PTB \cite{marcus1993building} and C4 \citep{raffel2020exploring,dodge2021case} test set. 

For zero-shot tasks, we evaluate on ARC-challenge, ARC-easy \cite{clark2018think}, HellaSwag \cite{zellers2019hellaswag}, PIQA \cite{bisk2020piqa}, WinoGrande \cite{sakaguchi2020winogrande}. We employ the lm-eval-harness framework \cite{lm_eval_harness_v041} to evaluate the pruned model performance on these tasks.\footnote{
The version of lm-eval-harness used in this paper is the same as SliceGPT  \citep{ashkboos2024slicegpt}.
}

\paragraph{Implementation details.}

We run all pruning experiments in bfloat16 on
a single NVIDIA A100 80GB GPU. As calibration data, we sample 512
sequences of length 512 from the training split of WikiText-2
\cite{merity2016pointer}. All gate scores are initialized to zero, and we use a fixed STE temperature $\tau=1.5$ throughout pruning. 
We optimize only the structural scores of FFN
channels and KV head groups using AdamW (learning rate
$1\times10^{-2}$, batch size 1, 4 epochs, no weight decay). For LLaMA-2-7B, the entire pruning procedure finishes in roughly 6 minutes (0.1 GPU-hours).

\paragraph{Baselines.}
We compare GRASPrune against several state-of-the-art structured pruning approaches under similar post-hoc pruning budgets, including LLM-Pruner~\citep{ma2023llm}, SliceGPT~\citep{ashkboos2024slicegpt}, and FLAP~\citep{an2024fluctuation}, all of which prune a pretrained model with limited additional optimization or lightweight calibration, without task-specific fine-tuning.
By contrast, heavy compensation pipelines that couple structured pruning with layer-wise least-squares reconstruction or extensive LoRA/KD fine-tuning (e.g., SlimLLM~\citep{pmlr-v267-guo25a}) operate in a substantially larger retraining-budget regime and are therefore complementary to our focus.

We compare against baselines with publicly available implementations under a matched evaluation protocol; Appendix~\ref{app:pipeline_requirements} summarizes the pipeline assumptions and requirements of the considered methods.

\subsection{Main Results}
\paragraph{Performance.}

Table~\ref{tab:llama2_7b_full} provides a detailed view of GRASPrune on LLaMA-2-7B from 0.8 to 0.4 parameter retention. At 0.8, GRASPrune is already competitive with the strongest baselines, achieving the lowest WikiText-2 perplexity and the best average downstream accuracy. As sparsity increases to 0.6 and 0.5, its advantage becomes more pronounced: GRASPrune consistently attains the best perplexities on WikiText-2, PTB, and C4, and yields the highest average score over the zero-shot benchmarks, even if some baselines occasionally win on individual tasks.

The 0.4 retention setting is a high-sparsity stress test. Here, baseline methods degrade sharply in language modeling, whereas GRASPrune remains substantially more stable, outperforming FLAP and exceeding LLM-Pruner by over an order of magnitude. Downstream performance also drops across the board at this ratio, but GRASPrune preserves non-trivial capability on several benchmarks. Table~\ref{tab:multi-model-main} further shows that these trends generalize across backbones: at the same 0.4 ratio, GRASPrune consistently reduces perplexity to the tens and achieves the best average accuracy, indicating that global budgeted gating enables substantially higher compression while retaining useful language modeling ability.

Appendix \ref{app:new_model} further shows that the same trend generalizes to newer model families, including LLaMA-3.1-8B, Qwen3-8B, and Qwen3-14B, where GRASPrune remains consistently competitive and usually outperforms FLAP in average downstream accuracy under matched pruning ratios.

\paragraph{Prefill peak memory and KV-cache scaling.}
Table~\ref{tab:mem_breakdown_compact} reports the weight memory, the KV-cache memory, and the measured peak CUDA memory during the prefill stage for pruned LLaMA-2-7B under different param ratios.
Our key observation is that pruning reduces how fast memory increases with the prefill length $T$.
In particular, as we shrink the global budget, both the KV-cache and the peak prefill memory decrease in a near-proportional manner for the same $T$, implying a smaller length-dependent memory component.
We attribute this effect primarily to KV head pruning: for fixed precision and batch size, the KV-cache size scales with the number of retained KV head groups across layers, and thus decreases as more KV head groups are removed.

\begin{figure}[t]
    \centering
    \includegraphics[width=0.95\linewidth]{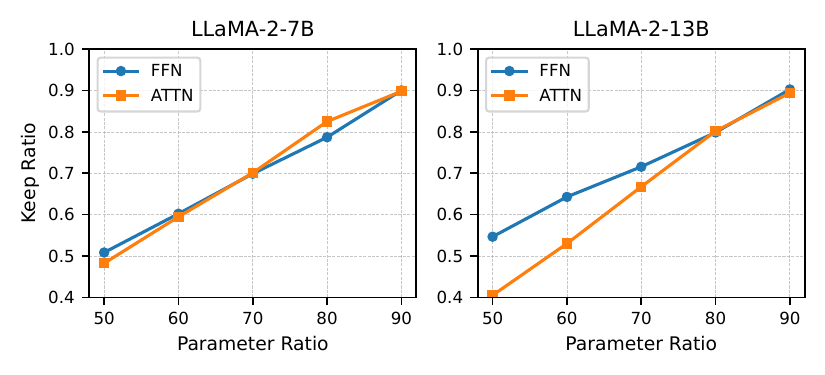}
    \caption{Comparison of global retention ratios between FFN channels and attention KV head groups under different sparsity targets on LLaMA-2-7B and LLaMA-2-13B.}
    \label{fig:retention per part}
\end{figure}

As an additional sanity check, Figure~\ref{fig:retention per part} shows that the retention of KV components remains relatively stable across param ratios, which helps explain the near-proportional reduction observed in KV-cache and peak prefill memory.
Overall, KV head pruning offers a practical deployment benefit by reducing the context-length-dependent memory growth, enabling longer prompts or larger batch sizes under the same GPU memory budget.

\paragraph{Throughput after structured pruning.}
\label{sec:throughput}

\begin{figure}[t]
  \centering
  \includegraphics[width=\linewidth]{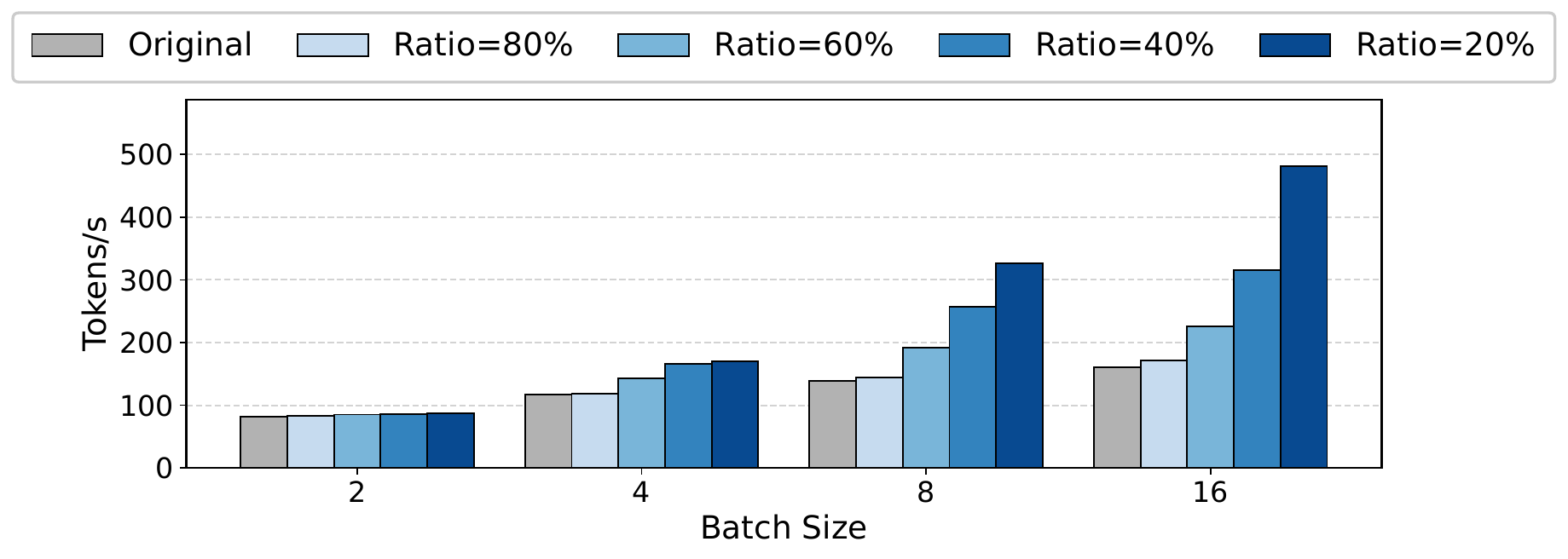}
  \caption{End-to-end generation throughput (tokens/s) of LLaMA-2-7B before and after pruning
  with prompt length 2048 and 256 generated tokens, measured under varying batch
  sizes. Ratios denote parameter retention.}
  \label{fig:throughput_batch}
\end{figure}

We report end-to-end generation throughput (tokens/s) of the dense LLaMA-2-7B model and our
materialized pruned checkpoints under identical inference settings, using a
fixed prompt length of 2048 tokens and generating 256 new tokens, while varying
the batch size.
Figure~\ref{fig:throughput_batch} shows that structured pruning consistently
improves throughput across batch sizes, with larger gains at higher batch sizes.
Moreover, more aggressive pruning yields higher
throughput, indicating that our pruned dense checkpoints provide practical
end-to-end serving benefits beyond parameter-count reductions.

\begin{figure}[t]
    \centering
    \includegraphics[width=1\linewidth]{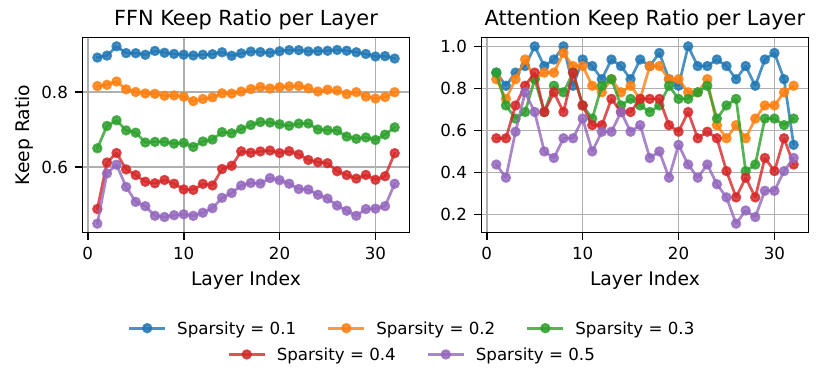}
    \caption{Layer-wise retention ratios of FFN channels
    and attention KV head groups under different global sparsity targets on LLaMA-2-7B.}
    \label{fig:retention per layer}
\end{figure}

\paragraph{Retention ratios.}
For each target budget, Figure~\ref{fig:retention per part} reports the resulting keep ratios of FFN channels and attention KV head groups.
As the budget tightens, both ratios decrease monotonically and remain broadly aligned, suggesting that the learned global allocation does not collapse to a single structure type.
Figure~\ref{fig:retention per layer} further breaks the allocation down by depth.
FFN retention changes smoothly across layers, whereas KV head retention shows larger layerwise variation under tighter budgets, with mid depth layers pruned more aggressively than the early and late blocks.

\subsection{Ablation Study}

\paragraph{Effect of the ranking criterion in the projection operator.}
As discussed in Section~\ref{sec:diff-mask}, our scores are learned under repeated budget-feasible projection. A natural alternative is to rank units by a cost-normalized criterion such as $p/c$. In our setting, however, replacing score ranking with $p/c$ changes the selection rule used after gate learning and empirically shifts the allocation toward cheaper units. Consistently, Table~\ref{tab:pc_ablation} shows that this modification degrades both perplexity and downstream accuracy. Appendix~\ref{app:pc_bias} further isolates this effect by fixing the learned utilities and changing only the ranking rule; under the same budget, the resulting allocation still skews strongly toward FFN, matching the observed quality drop.

\begin{table}[t]
\centering
\small
\setlength{\tabcolsep}{4pt}

\begin{tabular}{c|ccc|c}
\toprule
\textbf{Ranking} & \textbf{Wiki} & \textbf{PTB} & \textbf{C4} & \textbf{Avg.} \\
\midrule
Score $p$        & 6.47 & 48.18 & 11.44  & 0.614 \\
Value-per-cost $p/c$ & 9.37 & 53.75 & 13.59 & 0.553 \\
\bottomrule
\end{tabular}
\caption{Comparison of the ranking criterion in the budget projection operator
on LLaMA-2-7B at 80\% parameter retention. We replace the default score-based
ranking by a naive value-per-cost heuristic, sorting candidates by $p_i/c_i$
instead of $p_i$ while keeping the same budget and calibration protocol.}
\label{tab:pc_ablation}
\end{table}

\paragraph{Effect of the scaling stage.}
We further ablate the effect of the lightweight scaling stage in Table~\ref{tab:scaling_ablation}.
At a mild pruning level (0.8 keep ratio), scaling slightly improves language modeling perplexity on all three corpora, while the average downstream zero-shot accuracy decreases marginally.
This suggests that when pruning damage is small, the extra affine calibration helps the pruned model better fit the calibration distribution in terms of token-level likelihood, but can also induce a tiny perturbation of the zero-shot decision boundaries.

In contrast, for moderate to high sparsity (0.6–0.4 param ratios), scaling improves downstream accuracy and improves most perplexity metrics overall. At the extreme 0.4 setting, scaling yields large perplexity reductions and still raises the downstream average from 0.393 to 0.406.
This pattern is consistent with our intuition that aggressive structured pruning causes substantial layer-wise distribution shifts, which a small scaling module can effectively “repair”. 
When the model is heavily compressed, this last-mile calibration recovers a non-trivial portion of the lost capacity at essentially no additional training cost.

\begin{table}[t]
\centering
\small
\setlength{\tabcolsep}{4pt}

\begin{tabular}{cc|ccc|c}
\toprule
\textbf{Param Ratio} & \textbf{Method} & \textbf{Wiki} & \textbf{PTB} & \textbf{C4} & \textbf{Avg.} \\
\midrule
\multirow{2}{*}{0.8} 
 & w/o scale & 6.94 & 63.83  & 11.58 & 0.621 \\
 & w/ scale  & 6.47 & 48.18  & 11.44 & 0.613 \\
\midrule
\multirow{2}{*}{0.6} 
 & w/o scale & 10.42 & 76.99 & 20.08 & 0.482 \\
 & w/ scale  & 9.63  & 70.10 & 18.86 & 0.502 \\
\midrule
\multirow{2}{*}{0.5} 
 & w/o scale & 13.37 & 87.24 & 27.45 & 0.442 \\
 & w/ scale  & 12.18 & 92.87 & 27.45 & 0.465 \\
\midrule
\multirow{2}{*}{0.4} 
 & w/o scale & 20.40 & 173.51 & 50.49 & 0.392 \\
 & w/ scale  & 16.65 & 148.41 & 43.19 & 0.406 \\
\bottomrule
\end{tabular}
\caption{Effect of the scaling stage at different param ratios.}
\label{tab:scaling_ablation}
\end{table}

\section{Conclusion}

In this paper, we present GRASPrune, a global budgeted structured pruning framework that jointly prunes FFN channels and attention KV head groups under a single parameter budget. GRASPrune learns lightweight scalar gates on an unlabeled calibration corpus while keeping all backbone weights frozen, and then applies a simple scaling calibration on the retained structures to obtain a smaller dense model without task-specific fine-tuning. Experiments on LLaMA-style models show that our method better preserves language-modeling performance than existing structured pruning baselines across a wide range of parameter ratios, with especially large gains in high-compression regimes. These findings indicate that global, budgeted gating is a simple and effective mechanism for compressing LLMs while largely maintaining core language-modeling capability.

\section*{Limitations}
While we extend the evaluation beyond the main LLaMA-style setting and include additional analysis of calibration sensitivity, several limitations remain. We do not yet provide systematic evaluation on instruction-following, dialogue safety, code, or multilingual settings under the same lightweight pruning protocol. In addition, our unified budget is instantiated with a parameter-count proxy and a fixed FFN--KV cost ratio; although we separately report memory and throughput metrics, this proxy may not be optimal for all hardware or deployment objectives. Finally, while the appendix includes results on newer model families, the current validation is still not exhaustive across larger scales and broader deployment settings.

\paragraph{Potential risks.}
By reducing serving memory and latency, structured pruning may lower the barrier to deploying large models at scale, which could also increase misuse in unsafe settings. We therefore do not recommend deploying the pruned models in safety-critical scenarios without additional evaluation and mitigation.

\section*{Acknowledgments}
Sponsored by NSFC 62472289, 62532007 and Guangdong Province Key Laboratory of Popular High Performance Computers 2017B030314073.
\bibliography{custom}

\appendix

\begin{table*}[t]
\centering
\small
\renewcommand{\arraystretch}{1.2}
\setlength{\tabcolsep}{1.5pt}

\begin{tabular}{ll|l|ccc|cccccc}
\toprule
\textbf{Model} & \textbf{Ratio} & \textbf{Method} & \textbf{Wiki} & \textbf{PTB} & \textbf{C4}
& \textbf{ARC-C} & \textbf{ARC-E} & \textbf{HellaSwag} & \textbf{PIQA} & \textbf{WinoGrande} & \textbf{Avg.} \\
\midrule

\multirow{6}{*}{LLaMA-3.1-8B}
& \multirow{2}{*}{0.8}
& FLAP      &9.5996 &15.0140 &18.5394 &0.3294 &0.5366 &0.5771 &0.7176 &0.6346 &0.5591  \\
& 
& GRASPrune &9.3406 &25.7903 &20.0855 &0.3643 &0.5577 &0.6487 &0.7236 &0.6235 &0.5836  \\
\cmidrule(lr){2-12}
& \multirow{2}{*}{0.6}
& FLAP      & 18.3596 &29.0535 &41.5361 &0.2466 &0.4032 &0.4122 &0.6028 &0.5659 &0.4461  \\
&
& GRASPrune & 15.1613 &67.9484 &38.1156 &0.2867 &0.4394 &0.4802 &0.6404 &0.5462 &0.4786  \\
\cmidrule(lr){2-12}
& \multirow{2}{*}{0.5}
& FLAP      &25.1929 &38.3396 &59.4976	&0.2270 &0.3830 &0.3760 &0.5713 &0.5493 &0.4213  \\
&
& GRASPrune &18.8686 &81.9615 &70.1054 &0.2654 &0.4268 &0.3966 &0.6121 &0.5343 &0.4470  \\
\midrule

\multirow{6}{*}{Qwen3-8B}
& \multirow{2}{*}{0.8}
& FLAP      & 13.5906 &22.5397	&24.2751 &0.3831 &0.6199	&0.5838	&0.6801	&0.6022	&0.5738 \\
&
& GRASPrune &10.2586 &21.7176	&20.4018	&0.4394	&0.6835	&0.6427	&0.7291	&0.6180	&0.6225 \\
\cmidrule(lr){2-12}
& \multirow{2}{*}{0.6}
& FLAP      &25.1929	&52.7120	&56.3313	&0.2440	&0.3342	&0.4138	&0.5365	&0.5375	&0.4132 \\
&
& GRASPrune &15.4001	&38.7158	&33.6369	&0.3430	&0.5261	&0.4845	&0.6621	&0.5556	&0.5143 \\
\cmidrule(lr){2-12}
& \multirow{2}{*}{0.5}
& FLAP      &50.5938	&205.2479	&141.6168	&0.2500	&0.3552	&0.3184	&0.5637	&0.5122	&0.3999 \\
&
& GRASPrune &21.0494	&54.5981	&48.1826	&0.2765	&0.4318	&0.4107	&0.6034	&0.5627	&0.4570 \\
\midrule

\multirow{6}{*}{Qwen3-14B}
& \multirow{2}{*}{0.8}
& FLAP      &10.0406	&15.7653	&18.1103	&0.4872	&0.7075	&0.6882	&0.7339	&0.6914	&0.6616 \\
&
& GRASPrune &9.0533	&18.2881	&18.0046	&0.5162	&0.7323	&0.7018	&0.7682	&0.6646	&0.6766\\
\cmidrule(lr){2-12}
& \multirow{2}{*}{0.6}
& FLAP      &14.4671	&29.2814	&32.2852	&0.3379	&0.5189	&0.5089	&0.6393	&0.6022	&0.5214 \\
&
& GRASPrune & 12.5692	&29.6844	&26.6090	&0.3541	&0.5795	&0.5653	&0.6877	&0.6164	&0.5606 \\
\cmidrule(lr){2-12}
& \multirow{2}{*}{0.5}
& FLAP      &18.795	&45.2634	&45.9762	&0.2986	&0.4377	&0.4354	&0.5996	&0.5533	&0.4649 \\
&
& GRASPrune &14.467	&36.9429	&34.7047	&0.3268	&0.5480	&0.4789	&0.6523	&0.5612	&0.5134 \\
\bottomrule
\end{tabular}
\caption{
Generalization to newer model families under matched pruning settings.
We compare GRASPrune with FLAP on LLaMA-3.1-8B, Qwen3-8B, and Qwen3-14B at three retention ratios.
}
\label{tab:new-family-results}
\end{table*}

\begin{table*}[t]
\centering
\small
\renewcommand{\arraystretch}{1.15}
\setlength{\tabcolsep}{3pt}

\begin{tabular}{l|ccc|cccccc}
\toprule
\textbf{Gate learning corpus} & \textbf{Wiki} & \textbf{PTB} & \textbf{C4}
& \textbf{ARC-C} & \textbf{ARC-E} & \textbf{HellaSwag} & \textbf{PIQA} & \textbf{WinoGrande} & \textbf{Avg.} \\
\midrule
WikiText-2 & 6.94 & 63.83 & 11.58 & 0.395 & 0.637 & 0.683 & 0.761 & 0.631 & 0.621 \\
PTB        & 12.77 & 11.99 & 12.97 & 0.352 & 0.628 & 0.659 & 0.724 & 0.609 & 0.594 \\
C4         & 8.64 & 43.87 & 9.34  & 0.381 & 0.612 & 0.677 & 0.751 & 0.619 & 0.608 \\
\bottomrule
\end{tabular}
\caption{
Sensitivity of gate learning to the calibration corpus on LLaMA-2-7B at ratio $=0.8$.
}
\label{tab:gate-calib-sensitivity}
\end{table*}

\begin{table*}[t]
\centering
\small
\renewcommand{\arraystretch}{1.15}
\setlength{\tabcolsep}{3pt}

\begin{tabular}{l|ccc|cccccc}
\toprule
\textbf{Scaling calibration corpus} & \textbf{Wiki} & \textbf{PTB} & \textbf{C4}
& \textbf{ARC-C} & \textbf{ARC-E} & \textbf{HellaSwag} & \textbf{PIQA} & \textbf{WinoGrande} & \textbf{Avg.} \\
\midrule
WikiText-2 & 6.47 & 48.18 & 11.44 & 0.385 & 0.635 & 0.675 & 0.741 & 0.635 & 0.614 \\
PTB        & 7.74 & 14.93 & 11.81 & 0.358 & 0.604 & 0.651 & 0.725 & 0.594 & 0.586 \\
C4         & 7.05 & 42.52 & 10.10 & 0.381 & 0.649 & 0.671 & 0.731 & 0.628 & 0.612 \\
\bottomrule
\end{tabular}
\caption{
Sensitivity of scaling calibration to the calibration corpus on LLaMA-2-7B at ratio $=0.8$.
}
\label{tab:scale-calib-sensitivity}
\end{table*}

\section{Generalization to newer model families}
\label{app:new_model}
To evaluate whether the proposed framework transfers beyond the model family used in the main paper, we additionally test GRASPrune on newer backbones, including LLaMA-3.1-8B \cite{grattafiori2024llama3herdmodels}, Qwen3-8B, and Qwen3-14B \cite{yang2025qwen3technicalreport}, under matched pruning settings. The main paper focuses on LLaMA and LLaMA-2 style backbones because they remain the most commonly used models in prior structured pruning studies and therefore provide the fairest basis for comparison with established baselines. For the newer families, we compare against FLAP only. This is because LLM-Pruner and SliceGPT are implemented for the standard multi-head attention setting, whereas the newer backbones considered here mainly adopt grouped-query attention, which makes direct comparison with those methods unavailable under a matched implementation setting. As shown in Table~\ref{tab:new-family-results}, GRASPrune remains consistently strong across model families and pruning ratios, which suggests that the proposed framework does not depend on LLaMA-2-specific structural assumptions.

\section{Sensitivity to the calibration corpus}
\label{app:corpus}

We study how the choice of calibration corpus affects GRASPrune. Since the method uses only a small unlabeled calibration set, we run a controlled cross-corpus study on LLaMA-2-7B at ratio $=0.8$, keeping the calibration token budget and sequence length fixed. We vary the calibration corpus among WikiText-2, PTB, and C4, and evaluate all resulting models on the same perplexity and downstream benchmarks.

\subsection{Sensitivity of gate learning to the calibration corpus}

Table~\ref{tab:gate-calib-sensitivity} shows that gate learning is moderately sensitive to the calibration corpus. The average downstream accuracy varies from 0.621 to 0.594 across WikiText-2, PTB, and C4, indicating that the learned structure ranking remains reasonably stable under a fixed calibration budget. Perplexity varies more across corpora, as expected from corpus-matching effects, but these changes do not translate proportionally to downstream accuracy. Overall, WikiText-2 and C4 are slightly more stable than PTB.

\subsection{Sensitivity of scaling calibration to the calibration corpus}

Table~\ref{tab:scale-calib-sensitivity} shows a similar pattern for scaling calibration. With the pruning mask fixed, changing the scaling corpus shifts the average downstream accuracy from 0.614 to 0.586. This indicates that the post-pruning re-alignment is reasonably robust and not tied to a single calibration distribution. As above, perplexity is more sensitive than downstream accuracy, and WikiText-2 and C4 remain slightly more stable than PTB. This supports our default choice of WikiText-2 in the main paper.

\section{Additional evaluation on complex capabilities}
\label{app:complex}

\begin{table*}[t]
\centering
\small
\setlength{\tabcolsep}{5pt}

\begin{tabular}{lccccccccc}
\toprule
Model & Ratio & Method & GSM8K (5-shot) & MMLU (8-shot) & BoolQ (0-shot) & SciQ (0-shot) \\
\midrule
\multirow{7}{*}{LLaMA-2-7B}
& 1.0 & Original      & 0.146 & 0.457 & 0.779 & 0.911 \\
\cmidrule(lr){2-7}
& \multirow{2}{*}{0.8}  & FLAP      & 0.000 & 0.294 & 0.556 & 0.878 \\
&  & GRASPrune & 0.000 & 0.255 & 0.661 & 0.708 \\

\cmidrule(lr){2-7}
& \multirow{2}{*}{0.6} & FLAP      & 0.000 & 0.231 & 0.519 & 0.426 \\
&  & GRASPrune & 0.000 & 0.246 & 0.638 & 0.632 \\

\cmidrule(lr){2-7}
& \multirow{2}{*}{0.5} & FLAP      & 0.000 & 0.231 & 0.468 & 0.371 \\
&  & GRASPrune & 0.000 & 0.255 & 0.620 & 0.636 \\

\midrule
\multirow{7}{*}{LLaMA-3.1-8B}
& 1.0 & Original      & 0.500 & 0.634 & 0.821 & 0.943 \\
\cmidrule(lr){2-7}
& \multirow{2}{*}{0.8}  & FLAP      & 0.000 & 0.343 & 0.641 & 0.865 \\
& & GRASPrune & 0.017 & 0.296 & 0.689 & 0.881 \\

\cmidrule(lr){2-7}
& \multirow{2}{*}{0.6}  & FLAP      & 0.000 & 0.231 & 0.614 & 0.700 \\
&  & GRASPrune & 0.002 & 0.254 & 0.616 & 0.741 \\

\cmidrule(lr){2-7}
& \multirow{2}{*}{0.5}  & FLAP      & 0.000 & 0.229 & 0.606 & 0.688 \\
&  & GRASPrune & 0.000 & 0.254 & 0.621 & 0.604 \\

\bottomrule
\end{tabular}
\caption{Additional evaluation on GSM8K, MMLU, BoolQ, and SciQ under different pruning ratios.}
\label{tab:complex-capabilities}
\end{table*}
To complement perplexity and standard zero-shot benchmarks, we further evaluate GRASPrune on capability-oriented tasks, including GSM8K\cite{cobbe2021trainingverifierssolvemath}, MMLU\cite{hendrycks2021measuring}, BoolQ\cite{clark-etal-2019-boolq}, and SciQ\cite{welbl-etal-2017-crowdsourcing}, and compare against FLAP on LLaMA-2-7B and LLaMA-3.1-8B at multiple pruning ratios.

Table~\ref{tab:complex-capabilities} shows that complex capabilities degrade rapidly under training-free structured pruning, especially for mid-sized models. In particular, GSM8K quickly drops to nearly zero after pruning, while MMLU also declines substantially, indicating that multi-step reasoning and higher-level knowledge are especially sensitive to structural compression. Under high compression regimes, these tasks approach a region where method differences are no longer stably reflected by GSM8K or MMLU alone.

By contrast, BoolQ and SciQ degrade more smoothly across pruning ratios, making them more informative for measuring the preservation of basic language understanding. On these tasks, GRASPrune outperforms FLAP at most retention ratios, indicating better preservation of usable model capacity under the same budget. Overall, these results are consistent with the main findings: under training-free structured pruning, complex reasoning degrades first, while basic understanding provides a clearer comparison signal under strong compression.

\section{Selection bias of cost-normalized ranking}
\label{app:pc_bias}

\begin{table*}[t]
\centering
\small
\renewcommand{\arraystretch}{1.15}
\setlength{\tabcolsep}{5pt}

\begin{tabular}{lcccccc}
\toprule
Method & FFN keep & KV keep & FFN budget & KV budget & $\mathbb{E}[p \mid \mathrm{FFN}]$ & $\mathbb{E}[p \mid \mathrm{KV}]$ \\
\midrule
Decoupled $(p)$ & 79.91\% & 80.18\% & 66.77\% & 33.23\% & 0.5463 & 0.5747 \\
Cost-normalized $(p/c)$ & 99.87\% & 39.94\% & 83.44\% & 16.56\% & 0.4720 & 0.9341 \\
\bottomrule
\end{tabular}
\caption{
Global allocation under heterogeneous costs with a fixed global budget.
We report the retained ratios (by count), the consumed budget share, and the mean utility of the selected units for each structure type.
}
\label{tab:pc_bias_global}
\end{table*}

This appendix provides a diagnostic comparison of two ranking rules under the same learned utilities and the same global budget. To isolate the effect of the ranking rule, we keep the learned utility scores fixed and only change how units are ordered when constructing the final budget-feasible hard mask.

\paragraph{Setup.}
We consider structured units of two types: FFN intermediate channels and attention KV head groups.
Let $p_i \in (0,1)$ denote the learned utility of unit $i$ (computed as a sigmoid of the gate score), and let $c_i>0$ denote its cost.
We use the same global budget as in the main experiments: the target budget is set to a fixed keep ratio (0.8) of the total prunable cost.
We adopt the same cost model as in the main method: FFN channels have unit cost $c_{\text{ffn}}=1$, and KV head groups have a larger unit cost $c_{\text{kv}}=\alpha$ derived from parameter footprint.
All results below are obtained under identical learned utilities; the only difference is the ranking rule used before truncation.

\paragraph{Compared ranking rules.}
We compare two ranking strategies for constructing a budget feasible hard mask:
\begin{itemize}[leftmargin=*]
\item \textbf{Decoupled ranking (by utility):} rank all units by $p_i$ in descending order, and use costs only to enforce feasibility by truncating the ranked list at the budget boundary.
\item \textbf{Cost normalized ranking:} rank all units by $p_i/c_i$ in descending order, and similarly truncate to satisfy the same global budget.
\end{itemize}

\paragraph{Global allocation shift under the same budget.}
Table~\ref{tab:pc_bias_global} summarizes the resulting allocation across structure types. Under the same budget, cost-normalized ranking allocates a larger fraction of the budget to FFN channels and retains substantially fewer KV head groups. This shows that changing the ranking rule alone can substantially alter the resulting allocation across structure types. Notably, although the few KV head groups kept by $p/c$ have high average utility, the overall KV capacity is still reduced because a larger share of the budget is absorbed by FFN units.

\paragraph{Layerwise distribution.}

\begin{figure}[t]
\centering
\includegraphics[width=\linewidth]{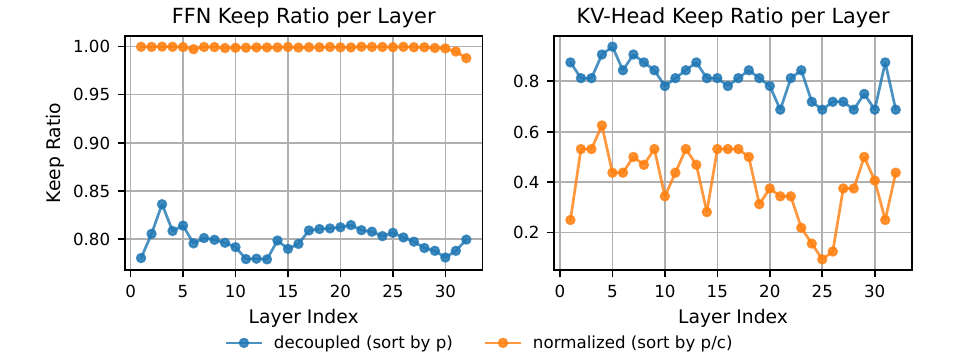}
\caption{Layerwise allocation under the same global budget.
We plot the per layer keep ratios of FFN channels (left) and KV head groups (right) for decoupled ranking (by $p$) versus cost normalized ranking (by $p/c$).}
\label{fig:pc_bias_layerwise}
\end{figure}

Figure~\ref{fig:pc_bias_layerwise} visualizes the per-layer keep ratios of FFN channels (left) and KV head groups (right) under the same global budget. Cost-normalized ranking preserves almost all FFN channels across layers while consistently shrinking KV capacity. In contrast, decoupled ranking retains fewer FFN channels but substantially more KV head groups in most layers, yielding a more balanced FFN/KV capacity split. Together with Table~\ref{tab:pc_bias_global}, this comparison shows that the ranking rule itself has a strong effect on the final allocation pattern under a fixed budget.

\section{Wall-clock overhead of the projection step.}
\label{app:wall_clock_proj}
The projection step involves sorting all prunable units and therefore has
an $O(N \log N)$ complexity, where $N$ is the total number of FFN channels
and KV head groups across layers. In practice, this cost is very small
compared with a Transformer forward/backward pass on GPU.

Under the hyper-parameters used in our experiments, on a single NVIDIA
A100 GPU with \texttt{torch.argsort}, the gate-learning stage for
LLaMA-2-7B performs 2048 sorts in total. The cumulative sorting time is
0.360\,s, while the total gate-learning time is 317.373\,s, meaning that
sorting accounts for only 0.11\% of the total time. For LLaMA-2-13B,
the gate-learning stage performs 6144 sorts in total, and sorting accounts
for 0.14\% of the total gate-learning time. These measurements suggest that
the sorting-based projection does not constitute a practical training
bottleneck.

\section{Sensitivity to the FFN--KV cost ratio.}
\label{sec:sensitivity_alpha}
Table~\ref{tab:alpha} reports the effect of perturbing the KV-head unit cost
by a scaling factor $s$ at 50\% parameter retention.
The best performance is obtained around the original estimate
$s{=}1$, but the method is reasonably robust to moderate misspecification:
for $s\in\{0.5,2\}$, perplexities change only slightly and the
average downstream accuracy varies within $2$--$3$ points.
Only more extreme distortions of the cost model ($s{=}4$ or $0.25$) lead to a
noticeable degradation (e.g., C4 perplexity increasing to 34.2 and the average
accuracy dropping to 0.433).
Overall, GRASPrune is not overly sensitive to the exact value of the FFN--KV
cost ratio, as long as it is specified within a reasonable range around the
true cost.

\begin{table}[t]
\centering
\small

\begin{tabular}{c|ccc|c}
\toprule
\textbf{Scaling $s$} & \textbf{Wiki} & \textbf{PTB} & \textbf{C4} & \textbf{Avg.} \\
\midrule
4.0  & 12.57 & 87.25 & 26.61 & 0.445 \\
2.0  & 13.59 & 79.44 & 29.68 & 0.437 \\
1.0  & 12.18 & 92.87 & 27.45 & 0.465 \\
0.5  & 12.77 & 95.82 & 29.22 & 0.457 \\
0.25 & 13.17 & 84.56 & 34.17 & 0.433 \\
\bottomrule
\end{tabular}
\caption{Sensitivity of GRASPrune to perturbations of the FFN--KV cost ratio
on LLaMA-2-7B at 50\% parameter retention. Each setting multiplies the
KV-head unit cost $\alpha_{\text{true}}$ by a factor $s$.}
\label{tab:alpha}
\end{table}

\section{Ablation on pruning targets}
\label{sec:ablation_targets}

\begin{table}[t]
\centering
\small

\begin{tabular}{l|ccc|c}
\toprule
\textbf{Pruning Target} & \textbf{Wiki} & \textbf{PTB} & \textbf{C4} & \textbf{Avg.} \\
\midrule
FFN-only  & 6.83  & 48.18  & 11.99  & 0.584 \\
KV-only   & 16.65 & 153.12 & 41.21  & 0.517 \\
Both      & 6.94  & 48.18  & 11.44  & 0.628 \\
\bottomrule
\end{tabular}
\caption{Ablation on pruning target under the same
global retention ratio (80\%) on LLaMA-2-7B. We report perplexity on WikiText-2,
PTB, and C4, and the average downstream accuracy.}
\label{tab:ablation_target_r08_llama2_7b}
\end{table}

At $\rho=0.8$ on LLaMA-2-7B, pruning KV head groups alone causes a large degradation in
language modeling perplexity, indicating that KV
capacity is substantially more sensitive in this mild-compression regime.
In contrast, FFN-only pruning maintains much lower perplexity, but yields a
lower downstream average than pruning both structure types jointly. This gap is
expected because restricting the search space to FFN channels forces the global
budget to be satisfied primarily by FFN reductions, preventing beneficial
trade-offs where the optimizer can allocate a small portion of the budget to KV
pruning while distributing the remaining reduction across FFN channels.
Overall, allowing FFN channels and KV head groups to compete under a single global
budget achieves a better perplexity--accuracy trade-off at the same retention
ratio.

\section{Mask stability during gate training}
\label{sec:mask_stability}

\begin{figure}[t]
    \centering
    \begin{subfigure}[ht]{0.49\linewidth}
        \centering
        \includegraphics[width=\linewidth]{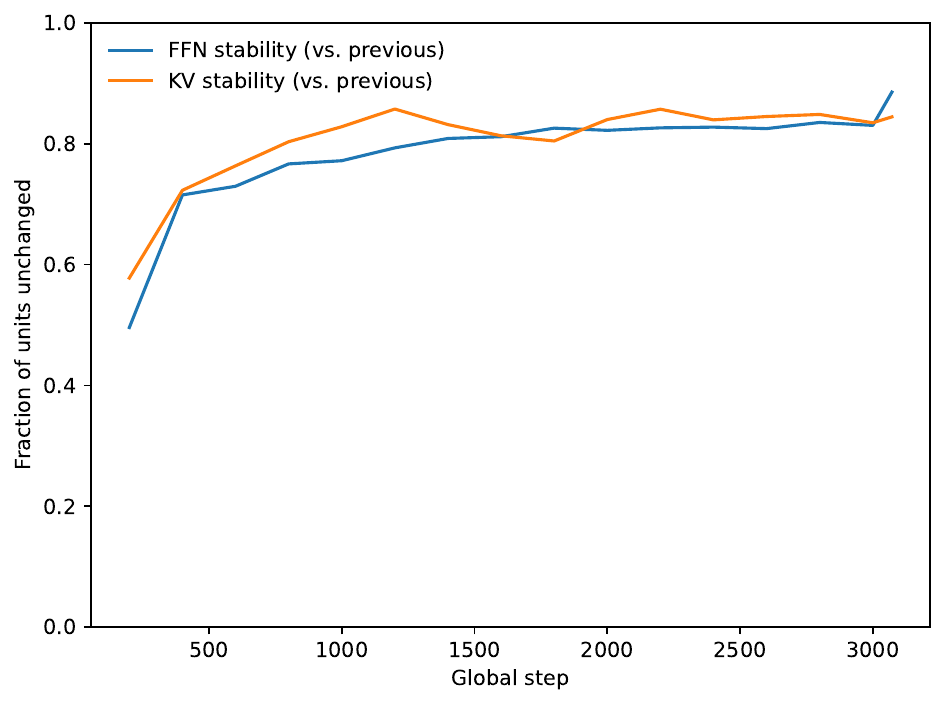}
        \caption{Budget-projected STE (ours).}
        \label{fig:mask_convergence_ste}
    \end{subfigure}
    \hfill
    \begin{subfigure}[ht]{0.49\linewidth}
        \centering
        \includegraphics[width=\linewidth]{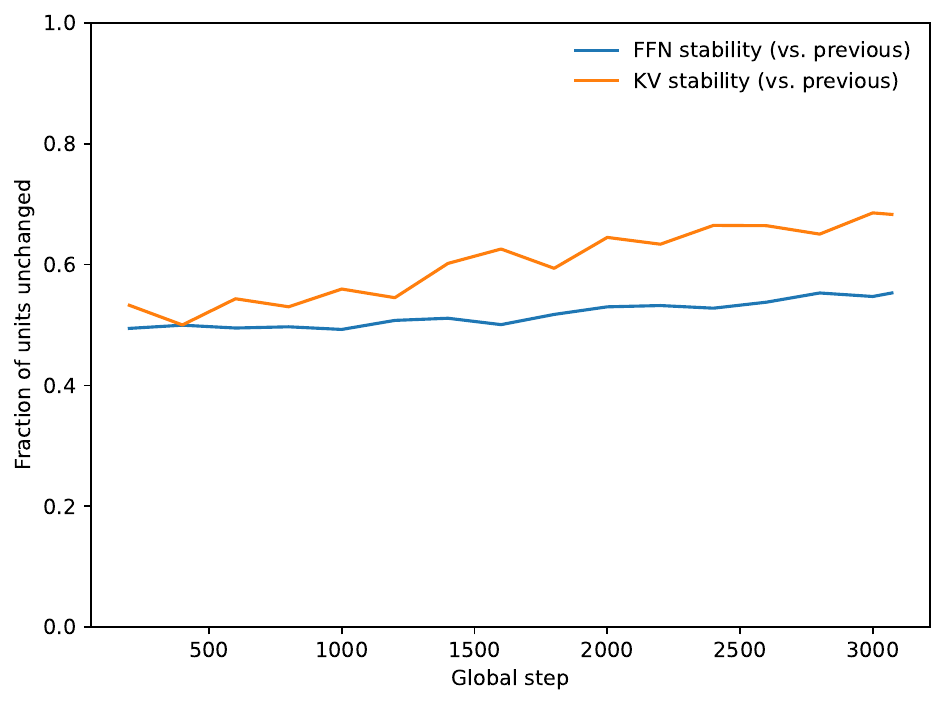}
        \caption{Gumbel-TopK gating baseline.}
        \label{fig:mask_convergence_gumbel}
    \end{subfigure}
    \caption{Mask stability comparison under identical settings (keep ratio 50\%, 512 calibration sequences, snapshot every 200 steps). We report the fraction of selected units whose on/off state remains unchanged between consecutive snapshots (kept vs.\ pruned), separately for FFN channels and KV head groups.}
    \label{fig:mask_convergence}
\end{figure}

We quantify the convergence behavior of our global, budget-feasible mask by tracking its stability during gate training.
Specifically, every 200 optimization steps we construct a discrete hard mask that exactly satisfies the global budget, and measure stability as the fraction of units that preserve their selection state between consecutive snapshots (kept vs.\ pruned), reported separately for FFN channels and KV head groups.

As shown in Figure~\ref{fig:mask_convergence_ste}, both FFN and KV stability rise quickly in the first few hundred steps and then plateau around 0.8--0.9 after roughly 1.5k updates, indicating that the learned global mask converges to a largely stable pattern while only a small subset of units near the budget boundary continues to be refined.
Residual fluctuations are expected because the gates are optimized with stochastic mini-batches and the forward pass applies a discrete, budgeted projection, so units with marginal scores may occasionally flip.
In practice, we find that 3--4 epochs are sufficient: longer training yields only marginal mask changes and negligible perplexity gains, but incurs additional pruning time.

\section{Global vs layer-wise budget allocation}
Table~\ref{tab:owl_vs_global} compares our global budget selection
with the layer-wise OWL allocation~\cite{yin2024outlier}.
Both methods share the same training protocol, differing only in how the pruning budget is
distributed.
At both 0.8 and 0.5 param ratios, the global scheme consistently
achieves lower perplexity on WikiText-2 for both LLaMA-7B and
LLaMA-2-7B, and the gap widens at higher sparsity.
This indicates that enforcing a single global budget, rather than prescribing a fixed per-layer sparsity schedule, allows the learned gates to reallocate capacity across layers and structure types based on their saliency.

\begin{table}[t]
\centering
\small
\setlength{\tabcolsep}{4pt}

\begin{tabular}{lcc|cc}
\toprule
\multirow{2}{*}{\textbf{Param Ratio}}
& \multicolumn{2}{c|}{\textbf{LLaMA-7B}}
& \multicolumn{2}{c}{\textbf{LLaMA-2-7B}} \\
\cmidrule(lr){2-3}\cmidrule(lr){4-5}
& \textbf{Global} & \textbf{OWL} & \textbf{Global} & \textbf{OWL} \\
\midrule
0.8 & 6.7278 & 6.7805 & 6.4700 & 6.8607 \\
0.5 & 11.9936 & 12.5692 & 12.1824 & 12.9681 \\
\bottomrule
\end{tabular}
\caption{Ablation on budget allocation. We compare our global budget
selection with a layer-wise OWL-style allocation on WikiText-2.}
\label{tab:owl_vs_global}
\end{table}

\section{Comparison to Gumbel-TopK gating}
\label{sec:compare_gumbel}

Our motivation for this comparison is the recent E$^3$-Pruner work~\cite{yuan2025e3prunerefficienteconomicaleffective}, which optimizes a discrete TopK pruning mask using a differentiable Gumbel-TopK sampler with a straight-through estimator (STE). Gumbel-TopK is a natural relaxation for layer pruning, where the candidate set is small and the mask dimension is low. However, its advantages are less clear in our setting, where the pruning decision is global and fine-grained, jointly selecting FFN channels and KV head groups across all layers, yielding a substantially higher-dimensional and more tightly budget-coupled decision space.

\begin{table*}[t]
\centering
\small
\setlength{\tabcolsep}{4pt}
\renewcommand{\arraystretch}{1.05}

\begin{tabular}{c r r r r}
\toprule
\textbf{Param Ratio} & \textbf{$T$} & \textbf{Weights (MiB)} & \textbf{KV-cache (MiB)} & \textbf{Peak (MiB)} \\
\midrule
1.0     &  256 & 12852.5 &  512.0 & 13990.8 \\
0.8 &  256 & 10382.1 &  418.5 & 11340.6 \\
0.6 &  256 &  7910.7 &  303.0 &  8644.5 \\
\midrule
1.0     &  512 & 12852.5 & 1024.0 & 15120.9 \\
0.8 &  512 & 10382.1 &  837.0 & 12260.0 \\
0.6 &  512 &  7910.7 &  606.0 &  9321.2 \\
\midrule
1.0     & 1024 & 12852.5 & 2048.0 & 17375.2 \\
0.8 & 1024 & 10382.1 & 1674.0 & 14114.8 \\
0.6 & 1024 &  7910.7 & 1212.0 & 10686.6 \\
\midrule
1.0     & 2048 & 12852.5 & 4096.0 & 21889.8 \\
0.8 & 2048 & 10382.1 & 3348.0 & 17822.5 \\
0.6 & 2048 &  7910.7 & 2424.0 & 13425.8 \\
\bottomrule
\end{tabular}
\caption{Full prefill-stage memory statistics for LLaMA-2-7B under different keep ratios (bfloat16, batch size $B=4$).}
\label{tab:mem_breakdown_full}
\end{table*}

\begin{table*}[ht]
\centering
\small
\setlength{\tabcolsep}{5pt}

\begin{tabular}{lccccc}
\toprule
\textbf{Method} &
\textbf{Teacher} &
\textbf{Solver-style} &
\textbf{Calibration} &
\textbf{Backprop} &
\textbf{Layer-wise} \\
& & \textbf{step} & \textbf{data} & & \textbf{processing} \\
\midrule
GRASPrune   & $\times$ & $\times$ & $\checkmark$ & $\checkmark$ & $\times$ \\
LLM-Pruner  & $\times$ & $\times$ & $\checkmark$ & $\checkmark$ & $\times$ \\
SliceGPT    & $\times$ & $\times$ & $\checkmark$ & $\times$     & $\checkmark$ \\
FLAP        & $\checkmark$ & $\times$ & $\checkmark$ & $\times$ & $\checkmark$ \\
OSSCAR      & $\times$ & $\checkmark$ & $\checkmark$ & $\times$ & $\checkmark$ \\
Olica       & $\times$ & $\checkmark$ & $\checkmark$ & $\times$ & $\checkmark$ \\
\bottomrule
\end{tabular}
\caption{
Pipeline requirements of representative post-training structured pruning methods.
\textbf{Teacher} indicates whether an additional teacher model is required beyond the pruned model itself (e.g., extra forward passes or targets from a separate checkpoint).
\textbf{Solver-style step} indicates whether the method includes an explicit optimization/solving procedure outside standard gradient backpropagation (e.g., combinatorial optimization, closed-form or iterative solvers).
\textbf{Calibration data} indicates whether a dedicated calibration set is needed.
\textbf{Backprop} indicates whether gradients are backpropagated through the pruned model during optimization.
\textbf{Layer-wise processing} indicates whether the method applies a per-layer reconstruction/solve pipeline (layers are processed largely independently), rather than a single global selection step.
This table summarizes algorithmic requirements rather than wall-clock runtime, which is implementation- and system-dependent.
}
\label{tab:pipeline_requirements}
\end{table*}

To isolate the effect of the gating mechanism, we implement a minimal Gumbel-TopK gating baseline under the same training protocol as our budget-projected STE gate training (same calibration data, keep ratio, optimizer, and snapshot frequency), and report the resulting mask stability curves in Figure~\ref{fig:mask_convergence_gumbel}. Compared to our method, Gumbel-TopK stabilizes noticeably more slowly and exhibits more frequent discrete flips, particularly near the budget boundary. This behavior is consistent with the intrinsic noise injection used by Gumbel-TopK: the sampler perturbs scores with Gumbel noise and forms a discrete TopK mask via a noise-driven relaxation, which increases the variance of effective hard decisions across optimization steps. In a high-dimensional joint selection space (FFN+KV), this stochasticity affects a large population of borderline units simultaneously, delaying the emergence of a stable global pattern. Notably, we observe an asymmetric effect across structures: KV head groups still show a gradual stabilization trend, whereas FFN channels remain considerably less stable, which we attribute to the much larger candidate pool and denser boundary region in the channel-wise selection space.

Beyond stability, the computational profile of Gumbel-TopK is also less favorable for fine-grained global gating. Unlike layer pruning where TopK is performed over tens of layers, our global mask spans orders of magnitude more candidates (all FFN channels and KV head groups), and the effective $k$ (kept units) is also large at moderate param ratios. In this regime, repeated sampling and relaxation steps introduce non-trivial overhead, making Gumbel-TopK less economical for cross-structure, budget-coupled pruning. Overall, Figure~\ref{fig:mask_convergence} suggests that budget-projected STE is better aligned with our problem setting: it reaches a stable, budget-feasible mask substantially faster while avoiding noise-induced oscillations in the joint FFN--KV selection space.

\section{Full memory statistics across prefill lengths}
\label{sec:mem_breakdown_full}

Table~\ref{tab:mem_breakdown_full} extends Table~\ref{tab:mem_breakdown_compact} to a wider range of prefill lengths, and the same scaling trends remain consistent across all $T$.
In addition, we report the residual “other” memory (i.e., peak minus weights and KV-cache), which captures transient activations and operator workspaces such as attention/MLP projections; this term also decreases approximately linearly with the keep ratio, indicating that non-cache activation memory scales proportionally with the pruned model size.

\section{Baseline pipeline requirements}
\label{app:pipeline_requirements}

Table~\ref{tab:pipeline_requirements} contrasts the algorithmic assumptions made by different post-training structured pruning pipelines. Compared with solver- or layer-wise reconstruction-based approaches, GRASPrune performs a single global, budget-feasible selection and optimizes only lightweight gate/scaling variables on a small calibration set, without requiring a separate teacher model. While GRASPrune uses backpropagation for gate optimization, it avoids heavy fine-tuning of the large backbone parameters and does not introduce per-layer solvers. Notably, reconstruction-based compensation is orthogonal to our global selection mechanism and can be applied on top of the pruned structure; we intentionally do not stack such compensation in the main pipeline to isolate the effect of budget-feasible global pruning on quality retention.

\begin{table*}[t]
\centering
\small
\begin{tabular}{lcccccc}
\toprule
\textbf{Model} & \textbf{\#Params (B)} & \textbf{\#Layers} & \textbf{Model Dim.} & \textbf{FFN Dim.} & \textbf{\#Heads} & \textbf{\#KV Heads} \\
\midrule
LLaMA-7B      & 7.0  & 32 & 4096 & 11008 & 32 & 32 \\
LLaMA-2-7B    & 7.0  & 32 & 4096 & 11008 & 32 & 32 \\
LLaMA-2-13B   & 13.0 & 40 & 5120 & 13824 & 40 & 40 \\
Vicuna-7B     & 7.0  & 32 & 4096 & 11008 & 32 & 32 \\
LLaMA-3.1-8B  & 8.0  & 32 & 4096 & 14336 & 32 & 8  \\
Qwen3-8B      & 8.2  & 36 & 4096 & 12288 & 32 & 8  \\
Qwen3-14B     & 14.8 & 40 & 5120 & 17408 & 40 & 8  \\
\bottomrule
\end{tabular}
\caption{Backbone models and their sizes used in this paper.}
\label{tab:backbones}
\end{table*}

\section{Backbone model sizes }
We summarize the backbone models used in our experiments in Table~\ref{tab:backbones}. These configurations determine the FFN channel counts and KV-head groups that define our pruning units and budgets.

\end{document}